\documentclass[10pt,twocolumn,letterpaper]{article}

\usepackage{iccv}      % To produce the REVIEW version
\definecolor{iccvblue}{rgb}{0.21,0.49,0.74}
\usepackage[pagebackref,breaklinks,colorlinks,allcolors=iccvblue]{hyperref}

\usepackage[utf8]{inputenc} % allow utf-8 input
\usepackage[T1]{fontenc}    % use 8-bit T1 fonts
\usepackage{hyperref}       % hyperlinks
\usepackage{url}            % simple URL typesetting
\usepackage{booktabs}       % professional-quality tables
\usepackage{amsfonts}       % blackboard iccv symbols
\usepackage{nicefrac}       % compact symbols for 1/2, etc.
\usepackage{microtype}      % microtypography
\usepackage{xcolor}         % colors
\usepackage{bbm}
\usepackage{dsfont}
\usepackage{soul}
\usepackage{graphicx}
\usepackage{booktabs}
\usepackage{xcolor}
\usepackage{tabularx}
\usepackage{xcolor, colortbl}
\definecolor{almond}{RGB}{239, 222, 205}

\newcolumntype{Y}{>{\centering\arraybackslash}X}
\usepackage{adjustbox}
\usepackage{rotating}
\usepackage{multirow}
\usepackage{multicol}
\usepackage{amsmath}
\usepackage{booktabs}
\usepackage{tabularx}
\usepackage{makecell}
\usepackage{adjustbox}
\usepackage{float}

\usepackage{wrapfig}
\usepackage{fontawesome5}

\usepackage{algorithm}
\usepackage{algpseudocode}
\usepackage{amsmath}
\usepackage{tcolorbox}

\def\paperID{8191} % *** Enter the Paper ID here
\def\confName{ICCV}
\def\confYear{2025}

\usepackage{acro}
\DeclareAcronym{reveal}{
  short = REVEAL,
  long = RElation-based Video rEpresentAtion Learning,
}
\DeclareAcronym{vlm}{
  short = VLM,
  long = Video Language Models,
}
\DeclareAcronym{videoqa}{
  short = VideoQA,
  long = Video-Question-Answering,
}
\DeclareAcronym{mm-nce}{
  short = MM-NCE,
  long = Many-to-Many Noise Contrastive Estimation,
}
\DeclareAcronym{mil-nce}{
    short = MIL-NCE,
    long = Multiple Instance Learning and Noise Contrastive Estimation 
}
\DeclareAcronym{llm}{
  short = LLM,
  long = Large Language Model,
}
\DeclareAcronym{clip}{
  short = CLIP,
  long =  Contrastive Language-Image Pre-training,
}
\DeclareAcronym{blip}{
  short = BLIP,
  long = Bootstrapping Language-Image Pre-training
}
\DeclareAcronym{lvlm}{
  short = LVLM,
  long = Large Vision-Language Model
}
\DeclareAcronym{cls}{
  short = CLS,
  long = Classification
}
\DeclareAcronym{ffn}{
  short = FFN,
  long = Feed Forward Network
}
\DeclareAcronym{mse}{
  short = MSE,
  long = Mean Squared Error
}

\title{REVEAL: Relation-based Video Representation Learning for Video-Question-Answering}

\author{Sofian Chaybouti$^{1,2}$ \hspace{2em} Walid Bousselham$^{1,2}$ \hspace{2em} Moritz Wolter$^{3}$ \hspace{2em} Hilde Kuehne$^{1,2,4}$ \\
{\small
$^1$Goethe University Frankfurt \quad
$^2$Tuebingen AI Center/University of Tuebingen \quad
$^3$University of Bonn \quad
$^4$MIT-IBM Watson AI Lab} \\
}

\begin{document}
\maketitle

\begin{abstract}

\ac{videoqa} comprises the capturing of complex visual relation changes over time, remaining a challenge even for advanced \ac{vlm}, i.a., because of the need to represent the visual content to a reasonably sized input for those models. 
To address this problem, we propose
\acf{reveal} a framework designed to capture visual relation information by encoding them into structured, decomposed representations. 
Specifically, inspired by spatiotemporal scene graphs, we propose to encode video sequences as sets of relation triplets in the form of (\textit{subject-predicate-object}) over time via their language embeddings. 
To this end, we extract explicit relations from video captions and introduce a 
\acf{mm-nce}
together with a Q-Former architecture to align an unordered set of video-derived queries with corresponding text-based relation descriptions.
At inference, the resulting Q-former produces an efficient token representation that can serve as input to a \ac{vlm} for 
\ac{videoqa}.

We evaluate the proposed framework on five challenging benchmarks: NeXT-QA, Intent-QA, STAR, VLEP, and TVQA. It shows that the resulting query-based video representation is able to outperform global alignment-based \acs{cls} or patch token representations and achieves competitive results against state-of-the-art models, particularly on tasks requiring temporal reasoning and relation comprehension. The code and models will be publicly released upon acceptance.
\end{abstract}    
\vspace{-2em}
\section{Introduction}
\label{sec:intro}

Videos capture rich sets of information, including the static visual information of a scene and the dynamic evolution of actors, objects, and their relationships over time. 
Understanding these complex spatiotemporal relations poses a significant challenge for current video understanding systems, as all those aspects need to be represented efficiently. 
One of the main tasks in this context is the problem of \ac{videoqa} \cite{wu2021star,xiao2021next,li2023mvbench,yu2023anetqa}. 
%\hkc{Are we mixing datasets and approaches here?} \scc{removed approaches only, maybe I should add more} \mwc{I think its okay.}
Approaches that do well here usually rely on pre-trained vision-language image backbones like \acs{clip} \cite{radford2021learning} and
\acs{blip}2~\cite{li2023blip}, processing videos by extracting frame representations and combining these with \acp{llm}~\cite{maaz2023video,ko2023large}.
However, these models struggle with object relations~\cite{yuksekgonul2022and, lin2024rethinking}, action detection~\cite{bansal2024videocon, wang2023paxion, lin2024rethinking, momeni2023verbs}, and compositional understanding~\cite{lin2024rethinking, bansal2024videocon}, issues that are exacerbated with temporal sequences. 
While recent works have shown that \acp{llm} can compensate those limitations via strong language priors~\cite{ko2023large,li2023mvbench,wang2024elysium,maaz2023video}, image- and video-language approaches still mostly rely on global video-text alignment representations to encode the video input.

To address this problem, we propose \acf{reveal}. This framework learns video representations by explicitly modeling the content as object relations over time via relation triplets in the form of (\textit{subject-predicate-object}). 
Our relation-based approach is inspired by prior work from video scene graphs context~\cite{cong2021spatial, ji2020action, rodin2024action, urooj2023learning}. 
However, scene graphs usually encode triplets via class indices, limiting the setting to close-ended and hand-annotated small-scale scenarios and hindering scalability. Inspired by this, REVEAL seeks to leverage this representation to learn general open-ended and web-supervised representations for video data.

To achieve this, we first leverage \acp{llm} to convert captions into one or more relation triplets, allowing us to source triplets at scale. 
The resulting triplets can be considered minimum viable sentences, allowing a standard text encoding, \textit{e.g.}, by a sentence encoder, resulting in one embedding representation per triplet and $J$ relation embeddings to describe a particular video. 
On the video side, we leverage a Q-Former architecture to encode the visual representation of one or more frames into a fixed set of vision queries. 
To train the Q-Former, we must match the fixed number of unordered vision queries to a variable number of unordered text triplet representations. 
To address this problem, we propose a Many-to-Many Noise Contrastive Estimation (\ac{mm-nce}) loss formulation, which aligns two sets of matching but unordered, incomplete sets, \textit{e.g.}, in our case, vision-based queries with corresponding text-based relation embeddings. 
Practically, \ac{mm-nce} maximizes the similarity between matched query-relation pairs while contrasting them against all unmatched pairs. 
This allows us to train the Q-former so that the resulting query tokens approximate the relation encodings of the video.
The resulting vision queries can then be used to fine-tune a standard \ac{vlm} architecture to address video-language-related tasks such as \ac{videoqa}.

We evaluate \ac{reveal} on five VideoQA datasets, NeXT-QA, Intent-QA, STAR, VLEP, and TVQA, demonstrating competitive performance compared to state-of-the-art methods. It shows that query-based representations, empowered by \ac{mm-nce}, are particularly effective at connecting video and text when adapting to \acp{llm}. 
Our analysis further reveals that initializing the relation encoder with a contrastively trained sentence embedder significantly enhances semantic alignment compared to alternatives like \acs{clip}’s text encoder.
We summarize the contributions of this work as follows:
\begin{itemize}
\item We propose a new encoding for web-based video learning by modeling relations in videos as target representation.
\item We propose a \ac{mm-nce} loss for contrastive learning over two sets of matching but unordered, incomplete sets.
\item We provide an extensive evaluation showing the efficacy of query-based representations and the role of \ac{mm-nce} in the context of state-of-the-art \ac{videoqa} architectures.
\end{itemize}
\section{Related Work}
\label{sec:related_work}

\paragraph{\acp{vlm} for \ac{videoqa}}
Video understanding, particularly \ac{videoqa}, has witnessed significant advancement with the emergence of \acp{llm} and Large Vision-Language Models.
Early approaches to VideoQA emerged in response to increasingly challenging benchmarks designed to test various aspects of video understanding. The complexity of VideoQA as a task is evidenced by the diverse set of benchmarks, each targeting different reasoning capabilities: TVQA~\cite{lei2018tvqa} challenged models with understanding TV show content requiring integration of visual cues and dialogue; STAR~\cite{wu2021star} focused on situated reasoning about object interactions in indoor environments; NextQA~\cite{xiao2021next} emphasized causal and temporal reasoning across everyday activities; IntentQA~\cite{li2023intentqa} specifically tested models' ability to understand human intentions and motivations behind observed actions; and VLEP~\cite{lei2020more} evaluated models' capacity to predict future events based on observed video content. 

In addressing these challenges, early approaches predominantly treated VideoQA as a classification task, where video and question features were fed into classification layers to select from a fixed set of answer choices~\cite{jang2017tgif, xu2017video, fan2019heterogeneous}. These methods typically employed CNN-RNN architectures, attention mechanisms, or memory networks to capture temporal dynamics, but their classification-based paradigm fundamentally limited their reasoning capabilities and prevented them from leveraging the generative power and world knowledge inherent in modern LLMs. Graph-based approaches like SHG-VQA~\cite{urooj2023learning} and VGT~\cite{xiao2022video} attempted to model explicit relations between objects but remained constrained by closed-vocabulary limitations, small-scale datasets, and the classification-based framework. These methods struggled with reasoning tasks due to a lack of semantic understanding.

Recent approaches have explored the direct application of VLMs to videos. IG-VLM~\cite{kim2024image} represents videos as image grids, while SLOWFAST-LLaVA~\cite{xu2024slowfast} employs multi-scale temporal pooling for feature extraction. While effective for general understanding, these methods often struggle with complex temporal reasoning, which \ac{reveal} addresses through explicit relation modeling.
Further, the success of instruction-tuning in image-LLM connections~\cite{hu2021lora, li2023m, beyer2024paligemma, liu2024improved, xiao2023florence} has inspired similar approaches for video understanding. Video-ChatGPT~\cite{maaz2023video}, VideoChat~\cite{li2023videochat}, and their successors VideoChat2~\cite{li2023mvbench} and VideoGPT+~\cite{maaz2024videogpt+} focus on video-conversation capabilities. Notable advances include Video-Llama~\cite{zhang2023video}'s multi-modal processing, Video-LLaVA~\cite{lin2023video}'s unified representation space, and MotionEpic~\cite{fei2024video}'s "Video-of-Thought" framework.
Llama-VQA~\cite{ko2023large} and Vamos~\cite{wang2023vamos} finetune adapters specifically for VideoQA. LLaVA-Next-Interleave~\cite{li2024llava}, MPLUG-OWL-3~\cite{ye2024mplug}, and LLaVA-One Vision~\cite{li2024llavaone} have further advanced instruction-tuning approaches with powerful vision backbones. 
%\hkc{Maybe we don't need this part - }
Finally, recent works have focused on unsupervised frame selection for this task, like Sevila, Vila, and LVNet~\cite{yu2024self, wang2024vila, park2024too}. These approaches use large vision backbones and Gumbel-Softmax~\cite{jang2016categorical} to discriminate frames, achieving strong results with a handful of frames. While orthogonal to REVEAL's relation-based approach, future work could combine these methods for more efficient videoQA. 
\vspace{-3mm}
\paragraph{Video-Language Pretraining}
Video-language pretraining has evolved significantly, with diverse architectural paradigms emerging to address the challenges of temporal modeling and multimodal alignment. Several key approaches have shaped this landscape: Q-former-based architectures like BLIP-2~\cite{junnan2022blip} and its video adaptations~\cite{hang2023videollama, lin2023video} use query-based cross-attention to bridge vision and language models; encoder-decoder frameworks like InternVideo~\cite{wang2022internvideo} and InternVideo2~\cite{wang2024internvideo2} combine masked video modeling with video-language contrastive learning; and unified architectures such as All-in-One~\cite{alex2022all} employ "token rolling" for efficient temporal modeling. Contrastive learning approaches have been particularly influential, with works like FrozenBiLM~\cite{bain2021frozen}, VideoCLIP~\cite{hu2021videoclip}, CLIP4Clip~\cite{huaishao2021clip4clip} and CLIP2Video~\cite{han2021clip2video}, establishing effective video-text alignment techniques. Temporal modeling has been addressed through hierarchical approaches in HiTeA~\cite{qinghao2022hitea} and HERO~\cite{linjie2020hero}, while UniVL~\cite{huaishao2020univl} pioneered joint understanding and generation objectives. Recent advances include VidL~\cite{cheng2023vindlu}, which presents a progressive recipe for video-language model construction, and MERLOT~\cite{zellers2022merlot}, which leverages YouTube transcripts for self-supervised learning.

\vspace{-0.5em}
\section{RElation-based Video rEpresentAtion Learning (REVEAL)} \label{method}
\vspace{-0.5em}

REVEAL is a framework designed to capture visual relation information in videos by encoding them into structured, decomposed representations. 
%Unlike traditional approaches that rely on global video-text alignment, REVEAL explicitly models videos as sets of relation triplets (subject-predicate-object).
%\scc{is this useful?}
This section is structured as follows: Sec.~\ref{subsec:RelationExtraction} describes the relation triplet sourcing from video captions, Sec.~\ref{subsec:Architecture} the overall architecture of REVEAL, Sec.~\ref{subsec:RelationModeling} details the relation modeling, Sec.~\ref{subsec:MM-NCEloss} describes the MM-NCE loss for aligning unordered sets of relations, and Sec.~\ref{subsec:slowfast_method} covers the implementation details.

\begin{figure}[ht!]
    \centering
    %\vspace{-1.5em}
    \includegraphics[width=\columnwidth]{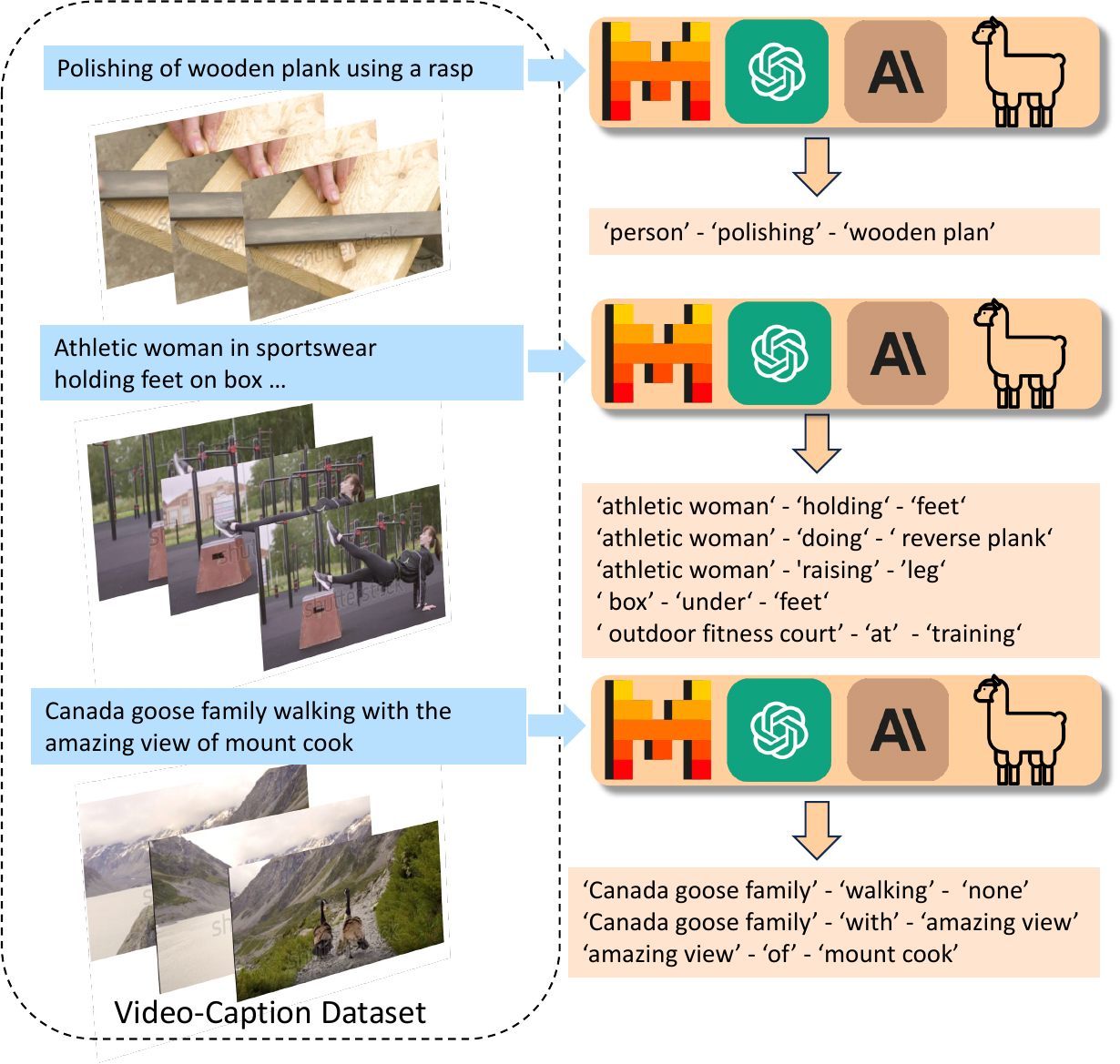}
    %\vspace{-2em}
    \caption{
        Relation extraction pipeline: Mistral-7B decomposes WebVid-2M captions into (subject-predicate-object) triplets.
    }
    \label{fig:annotation_pipeline}
\end{figure}

\begin{figure*}[t]
    \centering
    \includegraphics[width=\textwidth]{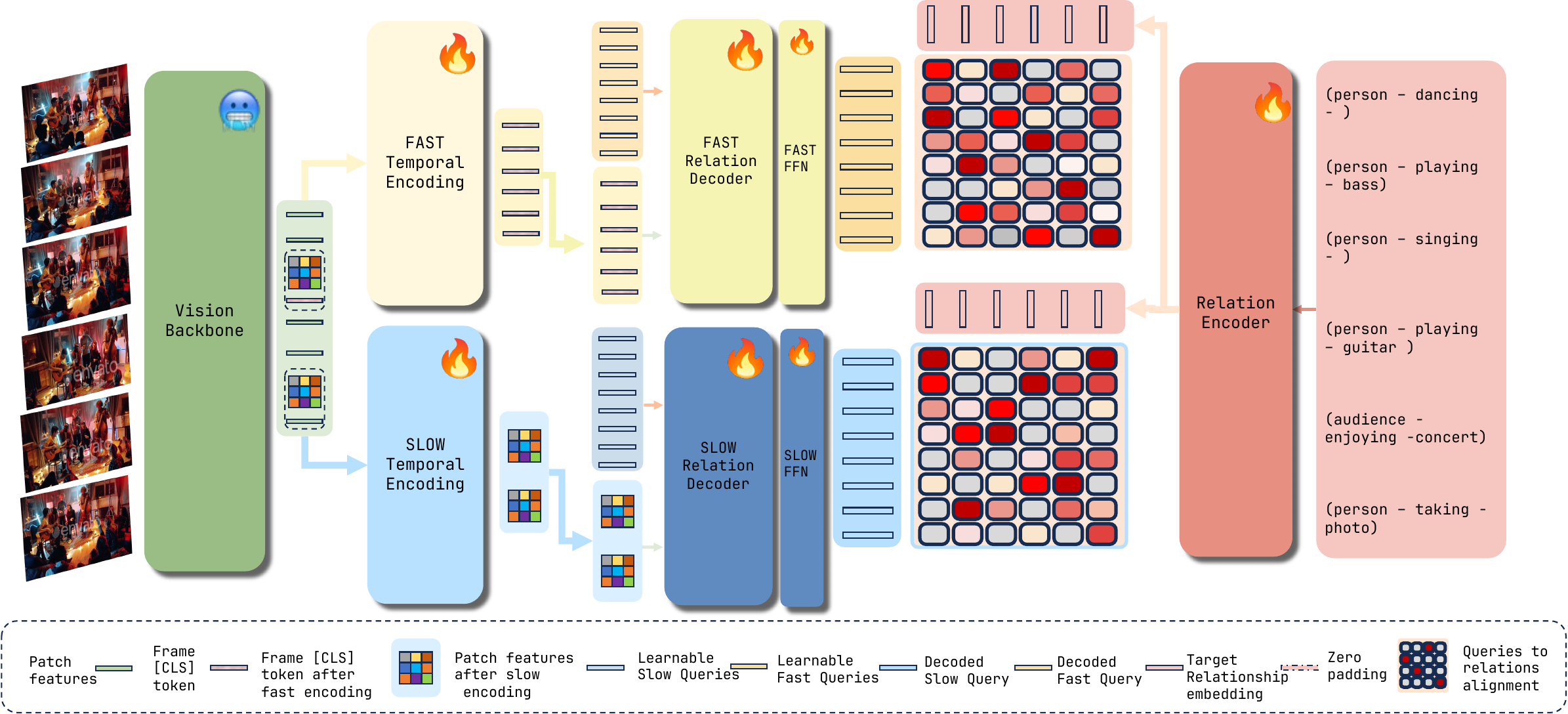}
    \vspace{-2em}
    \caption{
    REVEAL architecture for relation-based video representation learning. The model processes videos through dual pathways: a Fast Pathway (16 frames) for global context and a Slow Pathway (4 frames) for spatial details. Key components include Vision Encoders (CLIP ViT), Temporal Encoders (transformers), Relation Q-formers, and a Relation Encoder (Sentence-RoBERTa). The training uses our MM-NCE loss to align vision queries with text-derived relation triplets.
}
    \label{fig:model_architecture}
\end{figure*}
\vspace{-1em}
\subsection{Relation Extraction from Video Captions}
\label{subsec:RelationExtraction}
We develop a relation extraction pipeline to transform natural language video captions into structured relation triplets (\textit{subject-predicate-object}). Traditional approaches often depend on manually annotated datasets or rule-based methods, limiting scalability~\cite{thomee2016yfcc100m, yu2023anetqa, shang2019annotating, sigurdsson2016hollywood}. Compared to that, REVEAL leverages the Mistral-7B model~\cite{jiang2023mistral} to automate and scale extraction from large-scale datasets like WebVid-2M~\cite{bain2021frozen}.
To guide the LLM in decomposing unstructured captions into meaningful relation triplets, we in-context learning, detailed in Supplement Sec.~\ref{supp:sec:PromptEngineering}, to identify and extract relevant relations. This pipeline automatically generates multiple relation triplets per video, as illustrated in Figure~\ref{fig:annotation_pipeline}, providing a decomposed representation of the video caption respective to the visual content.
\vspace{-0.5em}
\subsection{REVEAL Architecture}
\vspace{-0.5em}
\label{subsec:Architecture}
Our approach represents videos as sets of relation triplets in the form of \textit{subject-predicate-object}. Unlike methods relying on finite indexed triplets~\cite{urooj2023learning} or separate object-predicate classification~\cite{herzig2023incorporating, salzmann2025scene}, we learn relation representations from language embeddings by aligning video-derived queries with text-derived relation embeddings.

As shown in Figure~\ref{fig:model_architecture}, the REVEAL architecture consists of four main components: (1) a vision encoder to compute frame-level features via a pretrained backbone; (2) a temporal encoder to capture the temporal dependencies across features from different frames; (3) a Relation Q-Former to transform the resulting visual features into vision queries; and (4) a Relation Encoder to encode text-based relation triplets for supervision.
During training, our MM-NCE loss aligns the vision queries with relation embeddings through Hungarian matching followed by contrastive learning, optimizing all components except the frozen vision backbone. 
\vspace{-1em}
\subsection{Relation Modeling}
\label{subsec:RelationModeling}

For a video $\mathcal{V}$, we begin with transforming the video into visual tokens. A visual encoder $f(.)$ processes each video's frames, producing a set of features: $(\mathbf{x}_n)_{n \in \{1..N\}} = f(\mathcal{V})$, where $N$ denotes the number of tokens per video. These tokens serve as input for relation modeling.

\textbf{Relation Q-former}: 
%
%To model relations from the visual tokens, we use a  Q-former module $g(.)$ that performs cross-attention between $M$ learnable queries $(\mathbf{v}^0_m)_{m \in \{1..M\}}$ and the visual tokens $(\mathbf{x}_n)_{n \in \{1..N\}}$, producing $M$ vision queries: $$
%(\mathbf{v}_m)_{m \in \{1..M\}} = g((\mathbf{v}^0_m)_{m \in \{1..M\}}, 
%%(\mathbf{x}_n)_{n \in \{1..N\}}),
%$$.
%
To transform learnable queries $(\mathbf{v}^0_m)_{m \in \{1..M\}}$ into vision queries $(\mathbf{v}_m)_{m \in \{1..M\}}$, we employ a Q-former architecture~\cite{detr}. 
This module performs cross-attention between the initial queries and the video's visual tokens $(\mathbf{x}_n)_{n \in \{1..N\}}$: $$
(\mathbf{v}_m)_{m \in \{1..M\}} = g((\mathbf{v}^0_m)_{m \in \{1..M\}}, (\mathbf{x}_n)_{n \in \{1..N\}}),
$$.
The resulting vision queries are processed through a feed-forward network to yield relation embeddings aligned with text-derived triplets.

\textbf{Relation Encoder}: In parallel, text relations $(\mathbf{t}_j)_{j \in \{1..J\}}$ associated with the video are passed through a text encoder $h(.)$ to get relation embeddings $(\mathbf{r}_j)_{j \in \{1..J\}} = h((\mathbf{t}_j)_{j \in \{1..J\}})$.
Practically, we leverage a pre-trained sentence embedder, initialized with contrastively trained models like Sentence-BERT~\cite{reimers-2019-sentence-bert}. %This initialization produces embeddings that are semantically rich and well-separated in vector space, enhancing discriminative power for distinct relations (e.g., "person opens door" vs. "person closes door").

Finally, the vision queries $(\mathbf{v}_m)_{m \in \{1..M\}}$ are aligned with the text-derived relation embeddings $(\mathbf{r}_j)_{j \in \{1..J\}}$ via the proposed \ac{mm-nce} loss. 
%This contrastive alignment trains the model to embed relations from the video input, leveraging caption-derived triplets as supervision.

\subsection{Relation Loss Function: Many-to-Many Noise Contrastive Estimation}
\label{subsec:MM-NCEloss}
We introduce Many-to-Many Noise Contrastive Estimation (MM-NCE) as a contrastive learning approach designed to align unordered sets of relations. The key challenge is that relation triplets extracted from video captions form an unordered set with no predefined temporal correspondence to visual elements in the video. This presents two difficulties: the number of extracted relation triplets may differ from the number of visual queries, requiring a flexible matching strategy, and unlike traditional video-text alignment where a single caption corresponds to an entire video, our approach must determine which specific vision query correspond to which relation embedding without explicit supervision. 
For a batch with samples $k \in \mathcal{B}$, we consider each video as $\mathcal{V}_k$, %and its respective set of relations as $\mathcal{J}^{(k)}=\{1..J^{(k)}\}$
the text-derived relation embeddings as $(\mathbf{r}^{(k)}_j)_{j \in \mathcal{J}^{(k)}}$, with $\mathcal{J}^{(k)} = \{1..J^{(k)}\}$, and the vision queries as $(\mathbf{v}^{(k)}_m)_{m \in \{1..M\}}$. We first determine the optimal matching between them using Hungarian matching, therefore creating a set of query-relation positive pairs for each video in the batch:
\begin{align}
\sigma^{(k)} = \text{argmax}_{\sigma \in \mathcal{S}_{J^{(k)}, M}} \sum_{j \in \mathcal{J}^{(k)}} s_c\left(\mathbf{r}^{(k)}_{j}, \mathbf{v}^{(k)}_{\sigma(j)}\right),
\label{eq:matching}
\end{align}
where $\mathcal{S}_{J^{(k)}, M}$ represents the set of injective mappings from $\mathcal{J}^{(k)}$ to $\{1..M\}$ and with the cosine similarity
\begin{align}
s_c(\mathbf{r}, \mathbf{v}) = \frac{\mathbf{r}^T \mathbf{v}}{\|\mathbf{r}\| \|\mathbf{v}\|}.
\end{align}
Note that, in equation \ref{eq:matching}, not all vision queries are paired to a text-derived relation embedding when $J^{(k)} < M$; the resulting mapping $\sigma^{(k)}(.)$ is injective but not surjective. This is a key property of our approach: it is designed to handle varying numbers of text relations per video. Eventually, only paired vision queries contribute to the loss defined below.

In the following equations, we omit the learnable parameters. The \ac{mm-nce} loss then consists of two symmetric terms.
$L_{\text{q}}$ measures query-to-relation alignment:
\begin{align}
L_{q \rightarrow r} = 
\sum\limits_{\substack{k \in \mathcal{B} \\ j \in \mathcal{J}^{(k)}}} 
\log \frac{\exp\left(s_c\left(\mathbf{r}^{(k)}_{j}, \mathbf{v}^{(k)}_{\sigma^{(k)}(j)}\right)/\tau\right)}
{\sum\limits_{\substack{k' \in \mathcal{B} \\ i \in \mathcal{J}^{(k')}}}
\exp\left(s_c\left(\mathbf{r}^{(k')}_{i}, \mathbf{v}^{(k)}_{\sigma^{(k)}(j)}\right)/\tau\right)},
\end{align}
and for the relation-to-query alignment term $L_{r \rightarrow q}$:
\begin{align}
L_{r \rightarrow q} = 
\sum\limits_{\substack{k \in \mathcal{B} \\ j \in \mathcal{J}^{(k)}}} 
\log \frac{\exp\left(s_c\left(\mathbf{r}^{(k)}_{j}, \mathbf{v}^{(k)}_{\sigma^{(k)}(j)}\right)/\tau\right)}
{\sum\limits_{\substack{k' \in \mathcal{B} \\ i \in \mathcal{J}^{(k')}}}
\exp\left(s_c\left(\mathbf{r}^{(k)}_{j}, \mathbf{v}^{(k')}_{i}\right)/\tau\right)}.
\end{align}

Here, $k'$ and $i$ index over all videos in batch $\mathcal{B}$ and vision queries from a video, respectively, creating negative pairs from other videos. The temperature parameter $\tau$ is learnable.

$L_{q \rightarrow r}$ and $L_{r \rightarrow q}$ allow us to compute
the \acs{mm-nce}-loss:
\begin{gather}
     L_{\text{MM-NCE}} = L_{q \rightarrow r} + L_{r \rightarrow q}.  
\end{gather}

\ac{mm-nce} pulls matched query-relation pairs closer in embedding space while pushing negative pairs apart, handling the unordered nature of relations through Hungarian matching rather than requiring predefined correspondences. It specifically allows the handling of varying numbers of annotations per video. When some vision queries are not matched to annotated relations, the model can freely learn to model relations in a video even when not annotated if they appear in other videos in the training data. Thus, it can also deal with non-exhaustive annotation. Unlike \ac{mil-nce}~\cite{miech20endtoend}, designed to align multiple captions to a single representation, this approach enforces a one-to-one correspondence between the multiple video representations, \textit{i.e.}, the vision queries, and the corresponding text relations.

% Compared to traditional video-text alignment~\cite{wang2023internvid, bain2021frozen, cheng2023vindlu} and closed-vocabulary methods~\cite{urooj2023learning, xiao2022video}, MM-NCE offers:
% \begin{itemize}
%     \item \textbf{Set-Based Matching}: Hungarian matching manages temporally and spatially unordered relations.
%     \item \textbf{Semantic Alignment}: The contrastively trained sentence embedder ensures embeddings align with natural language, enhancing open-vocabulary modeling.
%     \item \textbf{Robustness to Noise}: The loss accommodates non-exhaustive and noisy annotations, making it practical for large-scale, caption-derived data.
% \end{itemize}

\subsection{Implementation Details}
\label{subsec:slowfast_method}
\begin{figure}[ht!]
    \centering
    \includegraphics[width=\columnwidth]{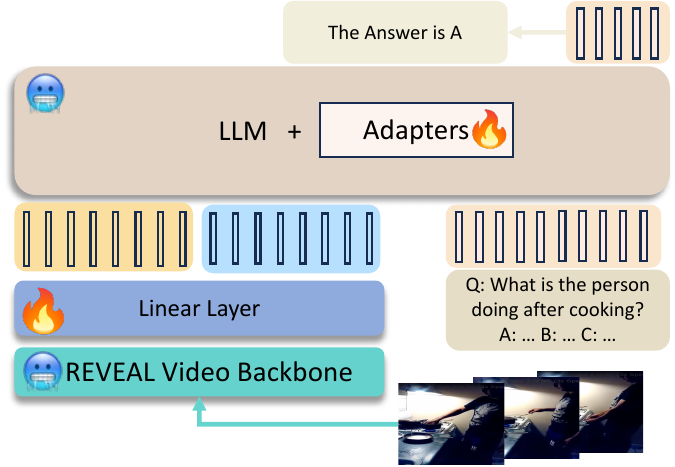}
    \caption{Overview of the VideoQA finetuning approach. The framework integrates pre-trained relation embeddings from our model with LLMs via adapters.}
    \label{fig:ft}
    \vspace{-1.5em}
\end{figure}

\paragraph{Slow-Fast Video Processing}
Following recent work~\cite{xu2024slowfast, maaz2024videogpt+}, \ac{reveal} employs a dual-pathway architecture to capture both global context and fine-grained spatial information, enhancing relation understanding while balancing computational efficiency: the \textbf{Fast Pathway} uses [\acs{cls}] tokens across 16 frames for efficient temporal aggregation and high-level motion understanding; the \textbf{Slow Pathway} processes patch features from four carefully selected frames for detailed spatial information and object-level relationship modeling.
Each pathway processes its respective features using a dedicated temporal encoder and relation Q-former. 
The temporal encoders model dependencies across frames, with the Fast pathway capturing global changes and long-range temporal dynamics and the Slow pathway specifically modeling patch relationships across frames for fine-grained spatial reasoning. 
The relation decoders perform cross-attention with visual features to transform learnable queries into relation embeddings representing meaningful subject-predicate-object triplets.
\paragraph{VideoQA Finetuning}
To evaluate REVEAL on multiple-choice VideoQA tasks, we adapt the frozen pre-trained video-derived relation features to LLMs, providing a decomposed representation for question answering, as illustrated in Figure~\ref{fig:ft}. Following previous work~\cite{wang2023vamos, ko2023large}, our finetuning approach integrates REVEAL's video relation embeddings into pre-trained LLMs using Llama adapters~\cite{zhang2023llama}. The process begins with REVEAL processing segmented videos (1–8 segments) in parallel, modeling 16 vision queries per segment (8 per pathway), which yields 16–128 embeddings per video. These embeddings are then projected into the LLM's vocabulary space via a linear transformation. For temporal alignment, each group of 16 vision queries corresponds to its respective video segment, with special tokens distinguishing between Slow and Fast Pathway outputs and learnable temporal tokens encoding segment positions. Our training methodology follows Flipped-VQA~\cite{ko2023large}, employing three complementary tasks: the main task (VQ→A) predicts answers from video vision queries and questions, while auxiliary tasks predict questions from vision queries and answers (VA→Q) and vision queries from questions and answers (QA→V). This multi-task approach reduces reliance on linguistic bias and enhances visual grounding, with REVEAL frozen to preserve the pre-trained representations.

\begin{table*}[htb!]
\vspace{-1em}
\centering
\resizebox{\textwidth}{!}{
    \begin{tabular}{l|c|l|l|c|c|c|c|c}
        \toprule
        \textbf{Method} & \textbf{Specifications} & \textbf{Language Backbone} & \textbf{Vision Backbone} & \textbf{Int} & \textbf{Seq} & \textbf{Pred} & \textbf{Feas} & \textbf{All} \\
        \midrule
        SHG-VQA (val set)~\cite{urooj2023learning} & FT & BERT & SlowR50-K400 & 48.0 & 42.0 & 35.3 & 32.5 & 39.5 \\
        All-in-One~\cite{wang2023all} & PT + FT & All-in-One & All-in-One & 47.5 & 50.8 & 47.7 & 44.0 & 47.5 \\
        InternVideo~\cite{wang2022internvideo} & PT + FT & CLIP text encoder & ViT-H/14 & 62.7 & 65.6 & 54.9 & 51.9 & 58.7 \\
        Sevila~\cite{yu2024self} & FT + FS & BLIP-2 (FlanT5-XL) & BLIP-2 (ViT-G/14) & \underline{63.7} & \textbf{70.4} & \underline{63.1} & \textbf{62.4} & \underline{64.9} \\
        ViLA~\cite{wang2024vila} & FT + FS & BLIP-2 (FlanT5-XL) & BLIP-2 (ViT-G/14) & \textbf{70.0} & \textbf{70.4} & \textbf{65.9} & \underline{62.2} & \textbf{67.1} \\
        \midrule
        IG-VLM~\cite{kim2024image} & ZS & Llava 1.6 & ViT-L/14 & 49.3 & 50.1 & 49.5 & 48.8 & 49.6 \\
        Llama-VQA~\cite{ko2023large} (baseline) & LLM-A & Llama1 & ViT-L/14 & \underline{66.2} & \underline{67.9} & \underline{57.2} & \underline{52.7} & \underline{65.4}  \\
        \rowcolor{almond} REVEAL (ours) & PT + LLM-A & Llama1 & ViT-L/14 & 60.0 & \underline{70.7} & \textbf{72.5} & \underline{68.4} & \textbf{67.9} \\
        \midrule
        Llama-VQA*~\cite{ko2023large} (baseline) & LLM-A & Llama3 & ViT-L/14 & \textbf{59.8} & \underline{67.2} & \underline{59.8} & \underline{50.4} & \underline{65.4} \\
        \rowcolor{almond} REVEAL (ours) & PT + LLM-A & Llama3 & ViT-L/14 & \underline{59.7} & \textbf{70.8} & \textbf{70.7} & \textbf{68.7} & \textbf{67.5} \\
        \bottomrule
    \end{tabular}
}
\vspace{-0.5em}
\caption{Performance comparison on STAR dataset for situated reasoning VideoQA across different question types (Interaction, Sequence, Prediction, and Feasibility). Specifications: PT = Pretraining, FT = Finetuning, FS = Frame Selection, ZS = Zero-Shot, LLM-A = LLM with Adapters. * indicates that we run the baseline evaluation ourselves.}
\label{tab:STAR}
\vspace{-1em}
\end{table*}
\section{Experiments} \label{experiments}
\vspace{-0.4em}
\begin{table*}
    \centering
    \resizebox{\textwidth}{!}{
        \begin{tabular}{l|c|l|l|c|c|c|c}
            \toprule
            \textbf{Method} & \textbf{Specifications} & \textbf{Language Backbone} & \textbf{Vision Backbone} & \textbf{Caus} & \textbf{Temp} & \textbf{Des} & \textbf{All} \\ 
            \midrule
            All-in-One~\cite{wang2023all} & PT + FT & All-in-One & All-in-One & 48.6 & 48.0 & 63.2 & 50.6 \\
            Video-Llama~\cite{zhang2023video} & IT + ZS & Llama & ViT-G/14 & 57.4 & 59.2 & 72.3 & 60.6 \\
            VideoChat~\cite{li2023mvbench} & IT + ZS & StableVicuna & BLIP-2 (ViT-G/14) & 61.5 & 63.5 & 82.1 & 61.8 \\
            HiTeA~\cite{ye2023hitea} & PT + FT & BERT-Base & MViT-Base & 58.3 & 62.4 & 75.6 & 63.1 \\ 
            InternVideo~\cite{wang2022internvideo} & PT + FT & CLIP text encoder & ViT-H & 58.5 & 62.5 & 75.8 & 63.2 \\
            VideoChat2~\cite{li2023mvbench} & IT + ZS & Llama1 & UMT-L & 64.7 & 68.7 & 76.1 & 68.6 \\
            LVNet~\cite{fei2024video} & ZS + FS &  GPT-4o &  GPT-4o & 65.5 & \textbf{75.0} &\textbf{81.5} & 72.9\\
            Sevila~\cite{yu2024self} & FT + FS & BLIP-2 (FlanT5-XL) & BLIP-2 (ViT-G/14) & \underline{69.4} & \underline{74.4} & 81.3 & \underline{73.8} \\
            ViLA~\cite{wang2024vila} & FT + FS & BLIP-2 (FlanT5-XL) & BLIP-2 (ViT-G/14) & \textbf{71.4} & 73.6 & \underline{81.4} & \textbf{74.8} \\
            \midrule
            IG-VLM~\cite{kim2024image} & VLM + ZS & LLava 1.6 & ViT-L/14 & 63.1 & 57.3 & 74.9 & 63.1 \\
            SLOWFAST-LLava~\cite{xu2024slowfast} & VLM + ZS & LLava-Next & ViT-L/14 & -- & -- & -- & 64.2 \\
            Video-ChatGPT~\cite{maaz2023video} & IT + ZS & LLaVA & ViT-L/14 & 64.1 & 66.9 & 75.7 & 64.4 \\
            Flipped-VQA (baseline)~\cite{kim2024image} & LLM + A & Llama1 & ViT-L/14 & 72.7 & 69.2 & 75.8 & 72.0 \\
            \rowcolor{almond} REVEAL (ours) & PT + LLM-A & Llama1 & ViT-L/14 & \underline{73.7} & \underline{69.2} & \underline{76.5} & \underline{72.7} \\
            \rowcolor{almond} REVEAL (ours) &  PT + LLM-A & Llama3 & ViT-L/14 & \textbf{75.3} & \textbf{69.9} & \textbf{78.5} & \textbf{74.0} \\
            \midrule
            Vamos~\cite{wang2023vamos}* & C + LLM-A & Llama3 & ViT-L/14 & 76.1 & 73.7 & 80.4 & 76.0 \\
            \rowcolor{almond} REVEAL (ours) & C + PT + LLM-A & Llama3 & ViT-L/14 & \textbf{77.8} & 74.4 & \textbf{81.9} & \underline{77.2} \\
            Vamos~\cite{wang2023vamos} & C + LLM-A & Llama3 & ViT-L/14 & \underline{77.2} & \underline{75.3} & \underline{81.7} & 77.3 \\
            \midrule
            LLaVA-Next-Interleave~\cite{li2024llava} & IT + ZS & QWEN-1.5 & SigLIP & -- & -- & -- & 77.9 \\
            MPLUG-OWL-3~\cite{ye2024mplug} & IT & QWEN-2 & SigLIP & -- & -- & -- & \underline{78.6} \\
            LLaVA-One Vision~\cite{li2024llavaone} & IT & QWEN-2 & SigLIP & -- & -- & -- & \textbf{79.4} \\
            \bottomrule
        \end{tabular}
    }

    \caption{Performance comparison on NExT-QA dataset for causal, temporal, and descriptive VideoQA. *indicates reproduced results.}
    \label{tab:NeXT-QA}
    \vspace{-2em}

\end{table*}
\subsection{Datasets}

\paragraph{Pretraining Datasets:}

We pretrain REVEAL on the \textbf{WebVid-2M} dataset, a large-scale collection of \textbf{2.5 million} video-caption pairs sourced from public web platforms~\cite{bain2021frozen}. 
%Spanning diverse content—tutorials, vlogs, sports, and entertainment—WebVid-2M provides comprehensive visual and contextual representations. 
Relation triplets are extracted from captions using the \textbf{Mistral-7B} model~\cite{jiang2023mistral}. 
Post-extraction, automated filtering removes ambiguous or redundant triplets, yielding an average of four relations per video.
To enhance relation diversity and robustness, we incorporate annotations of 8k videos from \textbf{Charades}~\cite{sigurdsson2016hollywood} and 3k videos from \textbf{VidOR}~\cite{shang2019annotating}, splitting each video into clips with 4–8 relations, adding approximately 80k clips to the training set. 
To prevent data leakage, we ensure no selected clips from Charades or VidOR overlap with STAR, NeXT-QA, or Intent-QA evaluation sets. The relation extraction and filtering details are provided in the supplement section \ref{supp:sec:PromptEngineering}.

\noindent\textbf{VideoQA Evaluation Datasets}:
We finetune and evaluate the resulting model on five diverse VideoQA benchmarks:
$\diamond$\textbf{STAR}~\cite{wu2021star} features 60K questions across four reasoning types (interaction, sequence, prediction, feasibility) with 22K indoor activity clips. Its procedurally generated questions require understanding object interactions and action consequences.
$\diamond$\textbf{NExT-QA}~\cite{xiao2021next} contains 52K manually annotated QA pairs over 5,440 videos, categorized as causal (48\%), temporal (29\%), or descriptive (23\%) testing event causation reasoning.
$\diamond$\textbf{Intent-QA}~\cite{li2023intentqa} extends NExT-QA with 16K QA pairs focused on intention understanding through four question types, challenging models to infer motives from observed actions.
$\diamond$\textbf{TVQA}~\cite{lei2018tvqa} comprises 152K QA pairs from 21K TV show clips, with five-choice questions. Its dialogue-heavy content requires integrating visual and linguistic cues across narratives. 
$\diamond$\textbf{VLEP}~\cite{lei2020more} presents a binary event prediction task with 28K examples across 10K clips. Models must predict which of two events will occur next, testing anticipatory reasoning.
Performance is measured by answer accuracy, with category breakdowns for STAR, NExT-QA, and Intent-QA highlighting task-specific strengths.

\begin{table*}[ht!]
\centering
\resizebox{\textwidth}{!}{
    \begin{tabular}{l|c|l|l|c|c|c|c}
        \toprule
        \textbf{Method} & \textbf{Specifications} & \textbf{Language Backbone} & \textbf{Vision Backbone} & \textbf{CW} & \textbf{CH} & \textbf{TP\&TN} & \textbf{All} \\ 
        \midrule
        HQGA~\cite{le2021hierarchical} & FT & BERT & ResNeXt-101/ResNet-101 & 48.2 & 54.3 & 41.7 & 47.7 \\ 
        VGT~\cite{xiao2022video} & FT & BERT & VGT & 51.4 & 55.9 & 47.6 & 51.3 \\  
        CaVIR~\cite{li2023intentqa} & FT & BERT & VGT & 58.4 & 65.4 & 50.5 & 57.6 \\ 
        VideoChat~\cite{lin2023video} & IT + ZS & StableVicuna & BLIP-2 (ViT-G/14) & -- & -- & -- & \underline{59.3} \\ 
        LVNet~\cite{fei2024video} & ZS + FS &  GPT-4o &  GPT-4o & \textbf{75.0} & \textbf{74.4} & \textbf{62.1} & \textbf{71.7} \\
        \midrule
        Video-LLaVA~\cite{lin2023video} & IT + ZS & Vicuna-7B & ViT-L/14 & -- & -- & -- & 62.5 \\ 
        Flipped-VQA*~\cite{ko2023large} & LLM-A & LLama3 & ViT-L/14 & \underline{73.7} & \underline{72.6} & \underline{57.3} & \underline{69.5} \\ 
        \rowcolor{almond} REVEAL (ours) & PT + LLM-A & Llama3 & ViT-L/14 & \textbf{74.0} & \textbf{77.4} & \textbf{66.8} & \textbf{72.8} \\
        \midrule
        Vamos~\cite{wang2023vamos} & C + LLM-A & Llama3 & ViT-L/14 & \underline{75.1} & \textbf{77.4} & \textbf{69.5} & \underline{74.1} \\ 
        \rowcolor{almond} REVEAL (ours) & PT + C + LLM-A & Llama3 & ViT-L/14 & \textbf{77.9} & \underline{77.3}& \underline{67.5} & \textbf{75.0} \\
        \bottomrule
    \end{tabular}
}
\vspace{-0.5em}
\caption{Performance comparison on Intent-QA dataset for intention understanding through causal and temporal reasoning (CW: Causal Why, CH: Causal How, TP\&TN: Temporal Previous \& Next). * indicates that we run the baseline evaluation ourselves.}
\label{tab:Intent-QA}
\vspace{-1em}
\end{table*}
\subsection{Training Details}
Using CLIP's ViT-L/14~\cite{radford2021learning} as the vision backbone, we process frame features with a two-layer transformer encoder followed by a 12-layer Q-former module for both the slow and the fast pathway output, resulting in eight learnable queries per pathway. 
This yields 16 vision query tokens per video clip, which are projected into the relation embedding space via a fully connected feed-forward network. %Orthogonal initialization of the queries enhances the diversity of vision queries, as validated in our experiments (see supplement).
The relation encoder is initialized with a pretrained sentence embedder ("all-roberta-large-v1" from~\cite{thakur-2020-AugSBERT}) based on Sentence-BERT~\cite{reimers-2019-sentence-bert} and the RoBERTa-large architecture~\cite{liu2019roberta}. 
It transforms relation triplets, formatted as "Subject: \textit{subj}, Predicate: \textit{pred}, Object: \textit{obj}", into single 1024-dimensional embeddings. 
Depending on the caption, one video can have multiple associated triplets. If more than eight text-derived relation embeddings are available, we randomly sample eight triplets.  
The resulting embedding sequence is further adapted with a one-layer feed-forward network.

We pretrain the model for five epochs on eight MI210 GPUs for approximately one day with the AdamW optimizer and a cosine-decayed learning rate of $5 \times 10^{-5}$. The resulting model comprises 590 million parameters. 

We finetune the model for each benchmark separately. 
To this end, we follow best practices of previous works~\cite{ko2023large, wang2023vamos}, keeping our pretrained video model, REVEAL, frozen and finetuning only a Linear layer and the Llama backbone via Llama-adapters~\cite{zhang2023llama} considering both Llama1 (7B) and Llama3 (8B)\cite{touvron2023llama, dubey2024llama} as our language models.
\vspace{-0.5em}
\subsection{Comparison to State-of-the-Art Methods}
We evaluate REVEAL against state-of-the-art methods across five VideoQA benchmarks.

\noindent \textbf{STAR:} Table~\ref{tab:STAR} shows the comparison with state-of-the-art approaches on STAR. We improve by $2.5\%$ compared to the Flipped-VQA baseline~\cite{ko2023large}, with the same vision, language backbones, and finetuning setting. Furthermore, we achieve state-of-the-art results improving upon ViT-G/14-based ViLA~\cite{wang2024vila} by $0.8\%$ while using the significantly less powerful ViT-L/14. The most substantial gains appear in prediction (+6.6\%) and feasibility (+6.2\%) questions testing the understanding of interactions and temporal reasoning. 

\noindent \textbf{NExT-QA:} On NExT-QA (Table~\ref{tab:NeXT-QA}), REVEAL with Llama3 achieves 74.0\% accuracy and 72.7\% with Llama1, outperforming the Flipped-VQA baseline (72.0\%) using identical vision backbones. 
Additionally, we implement a REVEAL+Captioning baseline for comparison with Vamos\cite{wang2023vamos}, integrating off-the-shelf captioning to complement relation embeddings with text descriptions.
This improves performance to 77.2\% and is on par with Vamos's results (77.3\%) while outperforming our reproduced baseline by 1.2\%.

\noindent \textbf{Intent-QA:} Table~\ref{tab:Intent-QA} shows REVEAL achieving 72.8\% accuracy on Intent-QA, beating the Flipped-VQA baseline 3.3\%. With complementary captions, REVEAL reaches 75.0\%, surpassing Vamos (74.1\%), with identical backbones. \\
%, demonstrating the effectiveness of relation-based representations for understanding motivations. \\
\textbf{TVQA and VLEP Datasets:} On TVQA (Table~\ref{tab:TVQA}), REVEAL achieves state-of-the-art performance (83.0\%), outperforming Flipped-VQA (82.2\%) by 0.8\%. Similarly, on VLEP (Table~\ref{tab:vlep}), we achieve 73.5\% surpassing Flipped-VQA by 2.5\% and 1.2\% with Llama1 and Llama3, respectively. 
These consistent improvements on datasets with different characteristics—from dialogue-heavy TV content to event prediction tasks—demonstrate the versatility and robustness of our relation-based approach.
\begin{table}[ht!]
\centering
\resizebox{\columnwidth}{!}{
\begin{tabular}{l|c|l|l|c}
    \toprule
    \textbf{Method} & \textbf{Specs.} & \textbf{Language} & \textbf{Vision} & \textbf{All} \\ 
    \midrule
    InternVid~\cite{wang2022internvideo} & PT+FT & CLIP & ViT-H & 57.2 \\
    Merlot~\cite{zellers2022merlot} & PT+FT & RoBERTa & ResNet-50 & \underline{78.7} \\
    VidL~\cite{cheng2023vindlu} & PT+FT & BERT & ViT-B/16 & \textbf{79.0} \\
    \midrule
    FrozenBiLM~\cite{bain2021frozen} & PT+ZS & DeBERTa & ViT-L/14 & 82.0 \\
    Flipped-VQA~\cite{kim2024image} & LLM-A & Llama1 & ViT-L/14 & \underline{82.2} \\
    \rowcolor{almond} REVEAL & PT+LLM-A & Llama3 & ViT-L/14 & \textbf{83.0} \\
    \bottomrule
\end{tabular}
}
\vspace{-0.5em}
\caption{Performance on TVQA dataset.}
\label{tab:TVQA}
\vspace{-1em}
\end{table}

\begin{table}[ht!]
\centering
\resizebox{\columnwidth}{!}{
\begin{tabular}{l|c|l|l|c}
    \toprule
    \textbf{Method} & \textbf{Specs.} & \textbf{Language} & \textbf{Vision} & \textbf{All} \\ 
    \midrule
    InternVideo~\cite{wang2022internvideo} & PT+FT & CLIP & ViT-H & \underline{63.9} \\ 
    Merlot~\cite{zellers2022merlot} & PT+FT & RoBERTa & ResNet-50 & \textbf{68.4} \\ 
    \midrule
    VideoChat~\cite{li2023mvbench} & IT+ZS & StableVicuna & ViT-G/14 & 62.0 \\ 
    SeViLA~\cite{yu2024self} & FT+FS & FlanT5-XL & ViT-G/14 & \underline{68.9} \\ 
    ViLA~\cite{wang2024vila} & FT+FS & FlanT5-XL & ViT-G/14 & \textbf{69.6} \\ 
    \midrule
    Video-LLaVA~\cite{lin2023video} & IT+ZS & Vicuna-7B & ViT-L/14 & 65.8 \\ 
    Flipped-VQA~\cite{kim2024image} & LLM-A & Llama1 & ViT-L/14 & 71.0 \\ 
    Flipped-VQA*~\cite{kim2024image} & LLM-A & Llama3 & ViT-L/14 & \underline{72.3} \\ 
    \rowcolor{almond} REVEAL & PT+LLM-A & Llama3 & ViT-L/14 & \textbf{73.5} \\
    \bottomrule
\end{tabular}
}
\vspace{-0.5em}
\caption{Performance on VLEP dataset. * indicates that we run the baseline evaluation ourselves.}
\vspace{-1.5em}
\label{tab:vlep}
\end{table}

\begin{table*}[th]
\centering
\resizebox{\textwidth}{!}{
\begin{tabular}{l|ccccc|cccc|cccc}
\toprule
\multirow{2}{*}{\textbf{Ablation}} & \multicolumn{5}{c|}{STAR} & \multicolumn{4}{c|}{NeXT-QA} & \multicolumn{4}{c}{Intent-QA} \\
& In & Seq & Pre & Feas & \textbf{All} & C & T & D & \textbf{All} & CW & CH & TN & \textbf{All} \\
\midrule
\rowcolor{almond}\textbf{\textit{a) Annotations:}} &  &  &  &  &  &  &  &  &  &  &  &  &  \\
Captions + NCE loss & 32.1 & 35.2 & 28.7 & 29.6 & 31.5 & 58.4 & 56.1 & 49.7 & 56.3 & 69.2 & 63.5 & 50.8 & 61.2 \\
relations + \ac{mm-nce} loss & \textbf{58.4} & \textbf{65.6} & \textbf{69.1} & \textbf{68.4} & \textbf{65.4} & \textbf{74.0} & \textbf{68.3} & \textbf{77.7} & \textbf{72.8} & \textbf{74.9} & \textbf{74.0} & \textbf{62.5} & \textbf{70.8} \\
\hline
\rowcolor{almond}\textbf{\textit{b) Rel. Enc.:}} &  &  &  &  &  &  &  &  &  &  &  &  &  \\
Frozen + MSE loss & 59.3 & 68.9 & 75.2 & 70.6 & 66.4 & 73.1 & 66.6 & 73.5 & 71.1 & 73.9 & 73.9 & 55.5 & 68.9\\
Frozen + \ac{mm-nce} loss & 59.7 & 69.0 & 73.1 & 69.8 & 67.9 & 73.8 & 68.8 & 76.2 & 72.6 & 73.7 & 74.4 & 63.1 & 71.4 \\
Trainable + \ac{mm-nce} loss & \textbf{61.4} & \textbf{69.3} & \textbf{75.0} & \textbf{72.0} & \textbf{69.4} & \textbf{75.3} & \textbf{69.9} & \textbf{78.5} & \textbf{74.0} & \textbf{74.6} & \textbf{75.5} & \textbf{65.6} & \textbf{71.8} \\
\hline
\rowcolor{almond}\textbf{\textit{c) LLM's video input:}} &  &  &  &  &  &  &  &  &  &  &  &  &  \\
Without FFN layer & 54.6 & 61.0 & 64.7 & 67.8 & 62.0 & 73.3 & 68.1 & 76.3 & 72.1 & 72.8 & 74.3 & 56.9 & 70.0 \\ 
With FFN layer & \textbf{61.4} & \textbf{69.3} & \textbf{75.0} & \textbf{72.0} & \textbf{69.4} & \textbf{75.3} & \textbf{69.9} & \textbf{78.5} & \textbf{74.0} & \textbf{74.6} & \textbf{75.5} & \textbf{65.6} & \textbf{71.8} \\
\hline
\rowcolor{almond}\textbf{\textit{d) Pathways:}} &  &  &  &  &  &  &  &  &  &  &  &  &  \\
Slow & \textbf{62.1} & \underline{68.9} & \underline{74.2} & \underline{70.2} & \underline{68.9} & 73.0 & 68.2 & 76.1 & 71.9 & 74.0 & 73.3 & 60.6 & 70.3 \\
Fast & 57.5 & 65.5 & 68.1 & 69.6 & 65.2 & \underline{73.7} & \underline{68.4} & \underline{77.5} & \underline{72.6} & \underline{73.1} & \underline{74.2} & \textbf{66.3} & \underline{71.1} \\
Slow-Fast & \underline{61.4} & \textbf{69.3} & \textbf{75.0} & \textbf{72.0} & \textbf{69.4} & \textbf{75.3} & \textbf{69.9} & \textbf{78.5} & \textbf{74.0} & \textbf{74.6} & \textbf{75.5} & \underline{65.6} & \textbf{71.8} \\
\bottomrule
\end{tabular}
}
    \caption{a) Pretraining on relations compared to training on captions. Both models were pre-trained on WebVid only. The caption model was contrastively trained by attention pooling on the vision queries. b) Ablation on the trainable relation encoder c) Results of using the vision queries compared to the last hidden states from REVEAL. d) Ablation on the slow-fast architecture.}
    \vspace{-1.5em}
\label{tab:abc}
\end{table*}

\begin{table}[ht]
\centering
\begin{tabular}{cccc}
\toprule
& STAR & NeXT-QA & Intent-QA \\
\midrule
\rowcolor{almond}\textbf{\textit{a) Initialization:}} &  &  &  \\
Random init & 68.3 & 71.0 & 69.2 \\ 
RoBERTa-large & 68.0 & 72.2 & 71.0 \\ 
CLIP text encoder & \underline{68.5} & \underline{72.3} & \underline{71.4} \\ 
Sentence embedder & \textbf{69.4} & \textbf{74.0} & \textbf{71.8} \\
\hline
\rowcolor{almond}\textbf{\textit{b) relations:}} &  &  &  \\
1 & 65.3 & 72.4 & 70.5 \\ 
2 & \underline{68.3} & 72.9 & 70.7 \\ 
4 & 68.0 & 73.1 & 71.3 \\ 
8 & \textbf{69.4} & \textbf{74.0} & \textbf{71.8} \\
\bottomrule
\end{tabular}
\caption{Ablation on a) the initialization of the relation encoder and b) the number of relations used as input to the LLM.}
\vspace{-2.em}
\label{tab:ab}
\end{table}

\subsection{Ablation Studies}
\vspace{-0.5em}
We conduct respective ablation studies to validate the key components of REVEAL. 
Results are summarized in Tables~\ref{tab:abc} and \ref{tab:ab} across STAR, NeXT-QA, and Intent-QA.

\noindent\textbf{Video-Relation vs. Video-Caption Alignment}: Table~\ref{tab:abc}.a tests relation modeling with \ac{mm-nce} loss compared to caption-based NCE supervision. 
It shows that pretraining with relations and \ac{mm-nce} loss yields significant improvements (STAR: +33.9\%, NeXT-QA: +16.5\%, Intent-QA: +9.6\%) over captions with standard NCE. 
These substantial gains validate the hypothesis that decomposing videos into structured relation triplets creates more effective representations than deriving representations from global captions.\\
\noindent\textbf{Trainable vs. Frozen Relation Encoder}: Table~\ref{tab:abc}.b provides first a baseline for the proposed matching loss, optimizing the query representation by computing the best matches and later optimizing them via MSE. Second, we provide results for the same setup but with the proposed \ac{mm-nce} loss function.
In both cases, the sentence embedder is kept frozen to evaluate the direct impact of the loss function, showing that \ac{mm-nce} provides better performance than matching followed by \acs{mse}. 
Third, we copy the second setup and make the sentence embedding trainable. 
A trainable encoder with \ac{mm-nce} loss consistently outperforms both a frozen encoder with \ac{mm-nce} (+1.5\% on STAR) and a frozen encoder with \acs{mse} loss (+3.0\% on STAR). 
This demonstrates that \ac{mm-nce} not only aligns relation sets but also enables the relation encoder to adapt to video-specific patterns, enhancing the semantic richness of our relation representations.\\
\noindent\textbf{Vision Queries vs. Hidden States}: We further evaluate the optimal input for the \ac{llm}. Namely, Table~\ref{tab:abc}.c compares the results for using tokens before and after the \acs{ffn} layer. 
This is motivated by the fact that the last layer in self-supervised learning can overfit on the objective. 
It shows that in our case, the output of the FFN projection outperforms the intermediate output of the Q-Former (STAR: +7.4\%, NeXT-QA: +1.2\%, Intent-QA: +1.8\%), confirming that explicitly modeling structured relations provides \acp{llm} with interpretable and actionable representations.\\
% 
%\begin{table}[htb!]
%\centering
%
%\begin{tabular}{@{}cl*{3}{c}@{}}
%\toprule
%Temp. Res. & STAR & NeXT-QA & Intent-QA & TVQA \\
%\midrule
%1 & 61.8 & \underline{72.9} & 70.7 & 79.6 \\
%2 & 65.1 & \textbf{74.0} & \textbf{71.8} & \underline{80.0} \\
%4 & \underline{66.7} & 72.6 & 70.5 & \textbf{80.3} \\
%8 & \textbf{69.4} & 72.7 & \underline{71.0} & {---} \\
%\bottomrule
%\end{tabular}
%
%\caption{Ablation on the temporal resolution. }
%\label{tab:temporal_res}
%\end{table}
\noindent\textbf{Slow-Fast Architecture}: Table~\ref{tab:abc}.d assess the impact of the dual-pathway architecture. It shows that using both representations consistently outperforms single-pathway variants (STAR: +0.5\% over Slow, +4.2\% over Fast), showing that modeling relations can be improved by detailed spatial information for object identification and efficient temporal modeling for action recognition.\\
\textbf{Relation Encoder Initialization}: Table~\ref{tab:ab}.a demonstrates that initializing the relation encoder with a contrastively trained sentence embedder significantly outperforms alternatives (e.g., +0.9\% over CLIP on STAR). This supports our claim that effective relation modeling requires semantically rich embeddings that can discriminate between similar but distinct relations (e.g., "person opens door" vs. "person closes door"), which contrastive training naturally provides.\\
\noindent\textbf{Number of Vision Queries}: Table~\ref{tab:ab}.b shows that increasing from 1 to 8 relations per pathway consistently improves performance (STAR: +4.1\%, NeXT-QA: +1.6\% and Intent-QA: +1.3\%), validating our modeling approach. This confirms that videos are better represented as sets of multiple relations rather than single global entities, with each additional relation contributing meaningful information.

\vspace{-0.5em}
\section{Conclusion}
\vspace{-0.5em}
We presented REVEAL, a framework advancing video understanding through relation-based representation learning. By modeling videos as relation triplet sets and introducing MM-NCE loss for aligning unordered relations, our approach creates structured embeddings that connect effectively with LLMs. Experiments show that decomposed relation-based representations outperform global alignment ones.
{
    \small
    \bibliographystyle{ieeenat_fullname}
    \bibliography{main}
}

\clearpage
\setcounter{page}{1}
\maketitlesupplementary

\printacronyms

\section{Pretraining Details}

\subsection{Dataloading}

Our data preprocessing and loading pipeline relies on WebDataset~\cite{webdataset_software}. Precomputed slow-fast CLIP features are stored as TAR files containing PyTorch~\cite{NEURIPS2019_9015} tensors. We use WebDataset's built-in shuffling mechanism with a buffer size of 5000 samples and an initial buffer of 1000 samples to ensure proper randomization.

\subsection{Model Implementation}
The pretraining model architecture consists of dual-pathway transformers processing slow and fast video features. We extract CLIP's patch features from the penultimate layer for the slow pathway. Each pathway includes a projection layer that maps 1024-dimensional input features to a hidden dimension $768$, followed by learnable positional encodings. The fast pathway processes CLS tokens features from 16 frames, while the slow pathway handles patch features from 4 frames. Both pathways utilize identical but separate transformer encoders, each comprising two encoder layers with 8-head self-attention (hidden size $768$, FFN dimension $4 \times 768$). The model employs fixed positional encodings using sinusoidal functions. We implement separate embedding modules for relationship modeling, generating $8$ learnable query embeddings for each pathway. The decoder architecture comprises 12 transformer decoder layers per pathway, each with 8-head cross-attention mechanisms and GELU activation functions. The decoder outputs are processed through an MLP with architecture $768 \rightarrow 4 \times 768 \rightarrow 1024$, where $1024$ is the ground truth embedding dimension. The implementation includes careful initialization strategies: orthogonal initialization for query embeddings, normal initialization (mean=0, std=0.02) for linear layers, and zero for biases. All normalization layers use LayerNorm.

\subsection{Pretraining}
Our pretraining implementation utilizes distributed training using PyTorch's DistributedDataParallel (DDP). The learning rate follows a cosine schedule with a linear warmup, starting from an initial learning rate of $lr = 0.00005$ with a 20\% warmup period over total steps, decaying to $0.05 \times lr$ at completion. Training proceeds for $5$ epochs with gradient accumulation every $4$ steps and gradient clipping at 1.0. We implement a bidirectional contrastive loss adapted to our multi-prediction setting following open-clip implementation \cite{cherti2023reproducible}. We use the Hungarian matching implementation from Scipy \cite{2020SciPy-NMeth} to match predictions with ground truth. The model employs two separate prediction heads for slow and fast pathways, each producing embeddings of dimension $1024$. We initialize the logit scale as $\log(1/0.07)$. We use the AdamW optimizer with a weight decay of 0.1. Training metrics are logged using Neptune.ai \cite{neptune2019}, including gradient norms, learning rates, and various losses. 

\section{Finetuning Details}

Our implementation leverages Llama-VQA implementation \cite{ko2023large}. Llama3 8B is the base language model, enhanced with REVEAL for video processing. We fine-tune using adapter layers while keeping the base Llama model frozen. Specifically, we use 32 adapter layers, with a length of tokens corresponding to the number of video relationships input to the LLM. The model extracts 16 relation queries per temporal segment, which are then linearly projected to match Llama's hidden dimension (4096). The training process uses AdamW optimizer with a base learning rate scaled by batch size (effective $lr = base\_lr \times batch\_size / 256$), with a linear warmup over 2 epochs and cosine decay. The training is done for five epochs. We use slow-fast features with a dimension of 1024 for video features, which are processed through REVEAL before being integrated with the language model. For datasets requiring subtitles (TVQA and VLEP), we integrate them into the input sequence before the question. All video frame features are pre-extracted and stored. In table \ref{tab:hpfinetuning}, we provide the hyperparameters per dataset.

\begin{table}[ht!]
\centering
\resizebox{\columnwidth}{!}{
\begin{tabular}{lccccc}
\toprule
\textbf{Hyperparameter} & \textbf{STAR} & \textbf{NextQA} & \textbf{Intent-QA} & \textbf{TVQA} & \textbf{VLEP} \\
\midrule
Base Learning Rate & 0.06 & 0.06 & 0.08 & 0.07 & 0.07 \\
Batch Size & 4 & 8 & 4 & 1 & 4 \\
Weight Decay & 0.14 & 0.1 & 0.14 & 0.02 & 0.12 \\
Temporal Resolution & 8 & 2 & 2 & 1 & 1 \\
Gradient Accum. & 8 & 4 & 4 & 4 & 2 \\
Bias & 3 & 3 & 3.5 & 3 & 3 \\
QAV loss & \checkmark & \checkmark & \checkmark & \checkmark & \checkmark\\
VAQ loss & \checkmark & \checkmark & \checkmark & \checkmark & $\times$\\
Max Sequence Length & 256 & 192 & 256 & 714 & 384 \\
\bottomrule
\end{tabular}
}
\caption{Dataset-specific hyperparameters used in our experiments. Values were determined through empirical validation.}
\label{tab:hpfinetuning}
\end{table}

\section{Full Ablation Tables}

Table \ref{tab:full_ablations} provides full per-category results for the temporal resolution, the relationship encoder initialization, and the number of relationships input to the LLM. The optimal temporal resolution, as expected intuitively, depends on the dataset. We also observe that the model with a relationship encoder initialized from a sentence embedder improves the performance of every question category evaluated. Finally, the more relationship vectors we input to the LLM, the better the results are, even though we get competitive results from a single relationship vector per temporal segment.

\begin{figure*}[ht!]
    \centering
    \includegraphics[width=\textwidth,height=0.9\textheight]{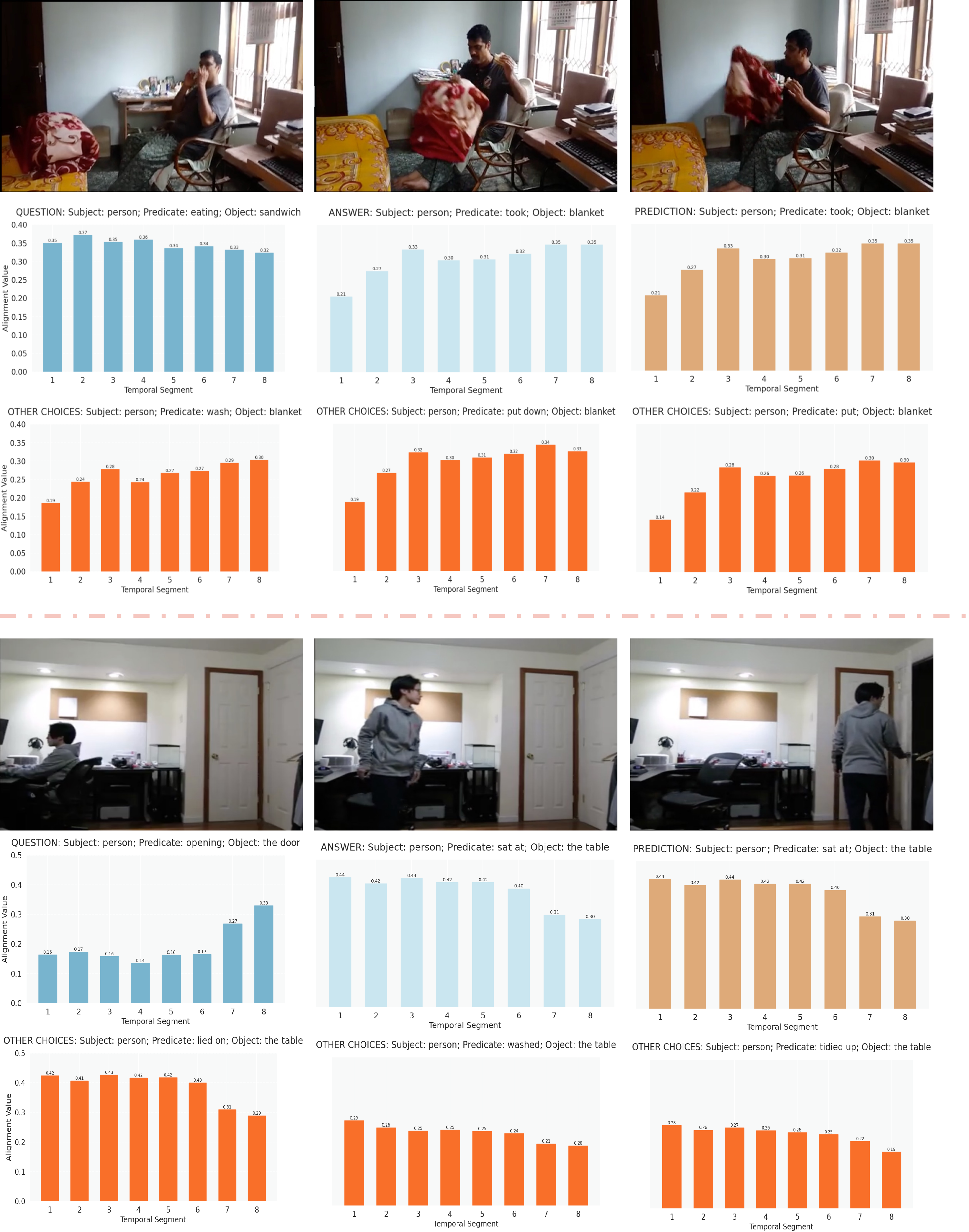}
    \caption{Successful examples from STAR dataset demonstrating REVEAL's relationship alignment capabilities. Top: The model correctly identifies concurrent actions (eating sandwich while taking blanket). Bottom: The model successfully captures temporal ordering of actions (sitting at table before opening door). Alignment scores between extracted relationships and video segments are visualized, showing stronger alignment during relevant temporal windows.}
    \label{fig:positive_examples}
\end{figure*}

\begin{figure*}[ht!]
    \centering
    \includegraphics[width=\textwidth,height=0.9\textheight]{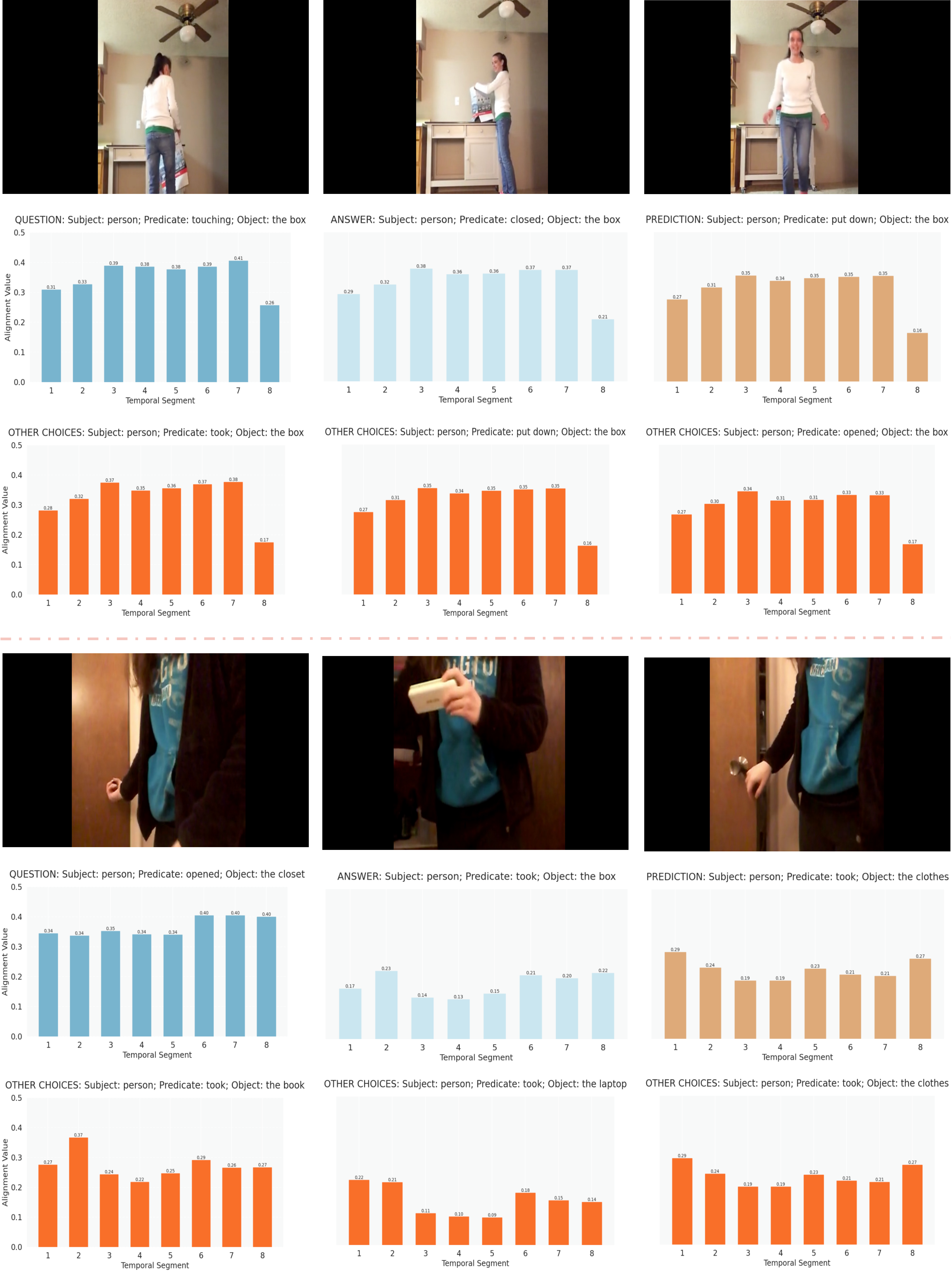}
    \caption{Failure cases from STAR dataset highlighting REVEAL's limitations. Top: Question ambiguity leads to multiple valid interpretations of the same action sequence. Bottom: Object recognition challenge where the model defaults to common-sense assumptions about closet contents rather than recognizing the specific object (small box).}
    \label{fig:negative_examples}
\end{figure*}

\section{Qualitative Analysis of the Performance on VideoQA}

\subsection{Successful Cases}
We present two successful examples from the STAR dataset where our model correctly answers the questions (Figure~\ref{fig:positive_examples}). In both cases, we visualize the alignment between the extracted relationship triplets and video segments (\textit{i.e.}, the maximum similarity scores between the decoded queries and the encoded relationships) to demonstrate how REVEAL processes temporal information.
In the first example, given the question "What is the person doing while eating a sandwich?", we extract the relationship triplet "Subject: person, Predicate: eating, Object: sandwich". The video is divided into 8 equal segments, and we observe strong alignment between this relationship and all segments, confirming the continuous eating action. The correct answer, "took blanket", shows increased alignment specifically during the relevant temporal window, while alternative choices exhibit lower alignment scores as these actions are absent in the video.
In the second example, for the question "What happened before the person opened the door?", we observe that the question's relationship becomes well-aligned with the video during the final two segments, corresponding to the door-opening action. The correct answer, "sat at the table", shows stronger alignment during the first six segments, maintaining higher scores than incorrect choices. 

\subsection{Failure Cases}
We also analyze two failure cases from the STAR dataset (Figure~\ref{fig:negative_examples}) to understand the model's limitations. The first case involves question ambiguity: given "What is the person doing after touching the box?", the model predicts "put down the box" while the ground truth is "closed the box". The alignment plots show that both relationships are well-matched with the video content, suggesting that both answers could be valid interpretations of the observed action sequence.
The second example ("What is the person doing after opening the closet?") demonstrates an object recognition challenge. While the correct answer involves taking a box, the model incorrectly predicts "take clothes". The alignment reveals the model's difficulty in recognizing the box, and the prediction may be influenced by Llama3's knowledge about items typically retrieved from closets.

\section{Prompt Engineering and Relationship Extraction}
\label{supp:sec:PromptEngineering}
\subsection{Prompt Template}

Figure \ref{fig:relationship_extraction} shows the complete prompt template used with Mistral-7B for relationship extraction. We leverage in-context learning by providing multiple examples of caption-relationship pairs before requesting the model to extract relationships from new captions. Each example demonstrates how to decompose a caption into subject-predicate-object triplets. The prompt includes diverse examples covering different types of actions, objects, and temporal relationships to encourage comprehensive extraction. This approach helps the model understand the expected format and granularity of the extracted relationships.

\subsection{Relationship Extraction Examples}
\begin{figure*}[ht!]
    \centering
    \includegraphics[width=\textwidth,height=0.9\textheight,keepaspectratio]{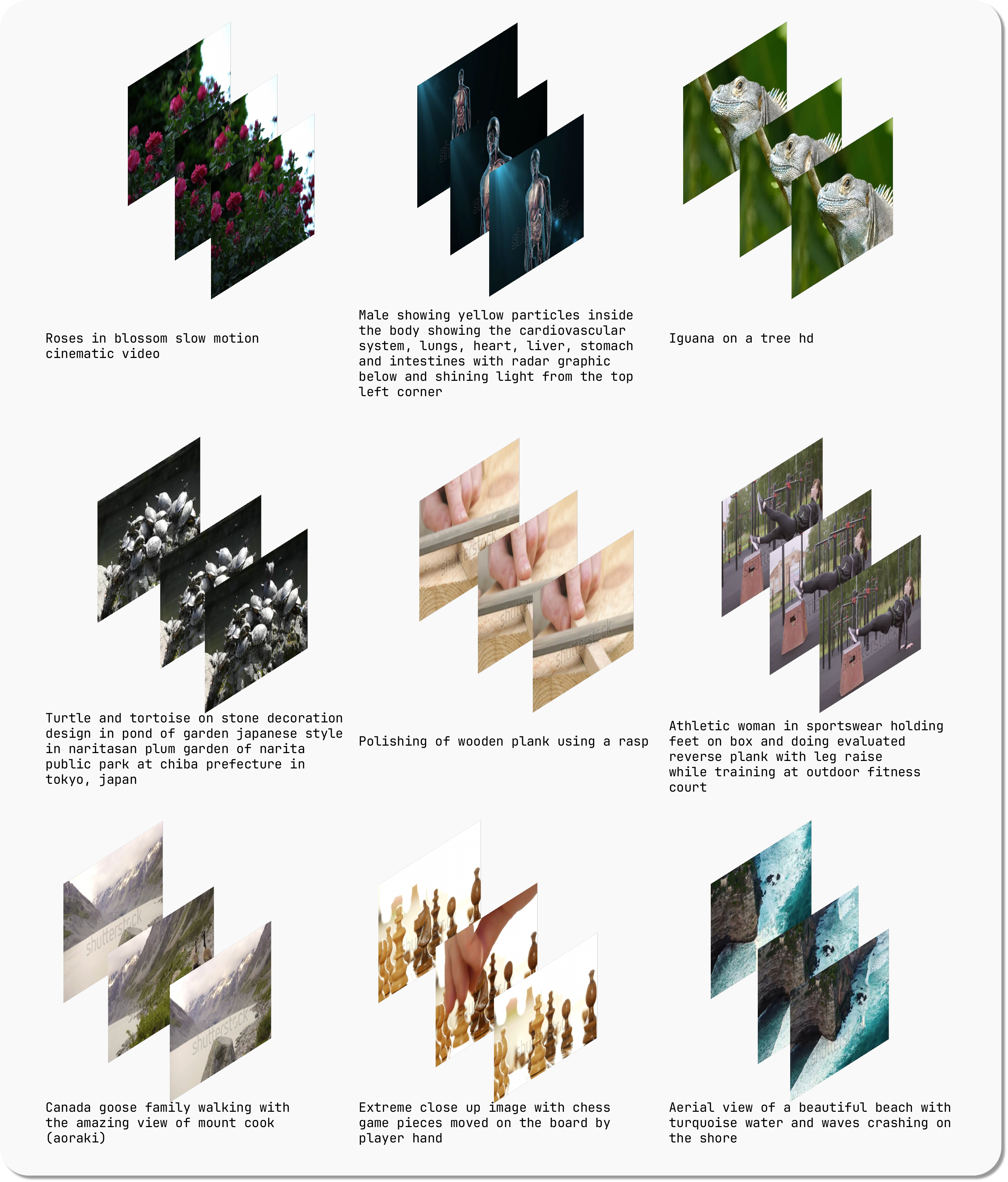}
    \caption{Sample videos from WebVid-2M}
    \label{fig:sample_webvid}
\end{figure*}

Table \ref{tab:relationships} presents examples of relationships extracted from Webvid-2M captions, with corresponding video frames shown in Figure \ref{fig:sample_webvid}. The extraction results demonstrate several key properties of our approach. The LLM generates a focused set of core relationships for concise captions. In contrast, complex or longer captions yield more detailed relationship sets. The extracted relationships, while accurate, are not exhaustive - they do not cover every possible relationship that could be inferred from the video content. This non-exhaustive nature of the extracted relationships validates our design choice not to penalize missing relationships during training to let the model freely infer relevant relationship vectors from videos.

\begin{table*}[t]
\resizebox{\textwidth}{!}{
\begin{tabular}{>{\centering\arraybackslash}m{0.4\textwidth}|p{0.56\textwidth}}
\toprule
\textbf{Caption} & \textbf{Relationships} \\
\midrule
Roses in blossom slow motion cinematic video & 
• Subject: roses , Predicate: in blossom, Object: none \\
& • Subject: roses , Predicate: appearing in, Object: cinematic video \\
& • Subject: cinematic video, Predicate: having, Object: slow motion \\
\hline
Male showing yellow particles inside the body & • Subject: male, Predicate: showing, Object: yellow particles \\
showing the cardiovascular system, lungs, heart, & • Subject: male, Predicate: showing, Object: body \\
liver, stomach and intestines with radar graphic & • Subject: body, Predicate: showing, Object: cardiovascular system \\
 below and shining light from the top left corner & • Subject: body, Predicate: showing, Object: lungs \\
& • Subject: body, Predicate: showing, Object: heart \\
& • Subject: body, Predicate: showing, Object: liver \\
& • Subject: body, Predicate: showing, Object: stomach \\
& • Subject: body, Predicate: showing, Object: intestines \\
& • Subject: radar graphic, Predicate: below, Object: male \\
\hline
Iguana on a tree hd & 
• Subject: iguana, Predicate: on, Object: tree \\
\hline
Turtle and tortoise on stone decoration design in & 
• Subject: turtle and tortoise, Predicate: on, Object: stone decoration \\
 pond of garden japanese style in naritasan plum & • Subject: turtle and tortoise, Predicate: in, Object: pond \\
garden of narita public park at chiba prefecture & • Subject: pond, Predicate: of, Object: garden \\
in tokyo, japan & • Subject: garden, Predicate: japanese style, Object: None \\
& • Subject: garden, Predicate: in, Object: Narita public park \\
& • Subject: Narita public park, Predicate: at, Object: Chiba prefecture \\
\hline
Polishing of wooden plank using a rasp & 
• Subject: person, Predicate: polishing, Object: wooden plank \\
\hline
Athletic woman in sportswear holding feet on box & 
• Subject: athletic woman, Predicate: holding, Object: feet \\
 and doing evaluated reverse plank with leg raise & • Subject: athletic woman, Predicate: doing, Object: reverse plank \\
while training at outdoor fitness court & • Subject: athletic woman, Predicate: raising, Object: leg \\
& • Subject: box, Predicate: under, Object: feet \\
& • Subject: outdoor fitness court, Predicate: at, Object: training \\
\hline
Canada goose family walking with the amazing & 
• Subject: Canada goose family, Predicate: walking \\
 view of mount cook (aoraki) & • Subject: Canada goose family, Predicate: with, Object: amazing view \\
& • Subject: amazing view, Predicate: of, Object: mount cook (aoraki) \\
\hline
Extreme close up image with chess game pieces & 
• Subject: player, Predicate: moving, Object: chess game pieces \\
 moved on the board by player hand & • Subject: player, Predicate: taking, Object: chess game pieces \\
& • Subject: chess game pieces, Predicate: on, Object: board \\
& • Subject: image, Predicate: close up \\
& • Subject: image, Predicate: containing, Object: chess game pieces and player hand \\
& • Subject: image, Predicate: having, Object: extreme close up perspective \\
\hline
Aerial view of a beautiful beach with turquoise& 
• Subject: view, Predicate: aerial, Object: beach \\
 water and waves crashing on the shore & • Subject: beach, Predicate: is, Object: beautiful \\
& • Subject: water, Predicate: is, Object: turquoise \\
& • Subject: waves, Predicate: crashing on, Object: shore \\
\bottomrule
\end{tabular}
}
\caption{Video Captions From Webvid-2M and Their Extracted Relationships}
\label{tab:relationships}
\end{table*}

\subsection{Relationship Extraction Pipeline for Charades and VidOR}

The VidOR and Charades datasets provide temporal annotations of relationships between objects in videos. Each relationship is annotated by a subject-predicate-object triplet and its temporal extent (start and end frames). To process these relationships into meaningful clips, we first collect all temporal ranges $(t_{start}, t_{end})$ for each video. We then employ a dynamic grouping algorithm that identifies natural breaks in the temporal annotations by analyzing the gaps between consecutive relationships. Specifically, we calculate the gap sizes between temporally adjacent relationships and use the 75th percentile of these gaps as a threshold to determine significant temporal breaks. This approach naturally segments the video into clips containing temporally coherent relationships. For each resulting clip, we aggregate all relationships whose temporal extent overlaps with the clip's timeframe, creating a set of relationships that describe the scene dynamics within that temporal window.

\begin{table*}[ht!]
\centering
\begin{tabular}{c|ccccc|cccc|cccc}
\toprule
\multirow{2}{*}{\textbf{Ablation}} & \multicolumn{5}{c|}{STAR} & \multicolumn{4}{c|}{NeXT-QA} & \multicolumn{4}{c}{Intent-QA} \\
& In & Seq & Pre & Feas & \textbf{All} & C & T & D & \textbf{All} & CW & CH & TP \& TN & \textbf{All} \\
\midrule
\rowcolor{almond}\textbf{\textit{a) Temp. Res.:}} & & & & & & & & & & & & & \\
1&54.9&62.3&64.1&65.9&61.8&74.2&68.8&77.0&73.4&74.3&61.0&55.0&70.7 \\
2&57.6&65.1&69.4&68.4&65.1&\textbf{74.0}&\textbf{70.0}&\textbf{77.9}&\textbf{73.3}&\textbf{74.6}&\textbf{75.5}&\textbf{65.6}&\textbf{71.8} \\
4&59.2&68.0&70.7&69.0&66.7&73.5&69.2&76.7&72.6&74.3&74.7&60.3&71.1 \\
8&\textbf{61.4}&\textbf{69.3}&\textbf{75.0}&\textbf{72.0}&\textbf{69.4}&73.7&69.6&75.4&72.7&72.2&75.0&60.6&70.8 \\
\hline
\rowcolor{almond}\textbf{\textit{b) Rel. Init.:}} & & & & & & & & & & & & & \\
Random& 59.7& 68.2& 72.9&72.5&68.3&72.4& 67.1& 74.4& 71.0& 70.7& 73.4& 58.7&69.2 \\
RoBERTa-large& 60.4& 67.9& 74.0& 69.6& 68.0& 73.8& 68.4& 74.9& 72.2&73.7&75.9&60.2&71.0 \\
CLIP text encoder& 59.4& 69.0& 73.1& 72.5& 68.5& 74.0&68.2&74.9&72.3&72.2&75.9&61.0&71.4 \\
Sentence embedder& \textbf{61.4}& \textbf{69.3}& \textbf{75.0}& \textbf{72.0}&\textbf{69.4}&\textbf{74.0}&\textbf{70.0}&\textbf{77.9}& \textbf{73.3}&\textbf{74.6 }&\textbf{75.5}&\textbf{65.6}&\textbf{71.8} \\
\hline
\rowcolor{almond}\textbf{\textit{c) \#Rels:}}& & & & & & & & & & & & & \\
1 & 57.8& 66.9& 68.4& 67.8& 65.3 & 73.8& 68.0& 77.2& 72.4 &72.5& 74.1& 60.8& 70.5 \\
2 & 62.1& 68.4& 74.0& 68.4& 68.3& 74.1& 68.9& 77.1& 72.9& 74.9& 74.3& 60.2& 70.7 \\
4 & 60.4& 67.7& 73.7& 70.2& 68.0& \textbf{75.1}& 68.3& 75.9& 73.1& \textbf{74.9}& 74.4& 61.8& 71.3 \\
8 & \textbf{61.4}&\textbf{ 69.3}& \textbf{75.0}& \textbf{72.0}& \textbf{69.4}& 74.0& \textbf{70.0}& \textbf{77.9}& \textbf{73.3}& 74.6& \textbf{75.5}&\textbf{ 65.6}& \textbf{71.8} \\
\bottomrule
\end{tabular}
%}
\caption{Full per category results for a) Temporal resolution; b) Relationship encoder initialization and; c) number of relationships vectors input to the LLM.}
\label{tab:full_ablations}
\end{table*}

\begin{figure*}[bh!]
\centering
\begin{tcolorbox}[
        colback=gray!5,      % background color (light gray)
        boxrule=1pt,         % border thickness
        arc=0pt,             % sharp corners
        width=\textwidth,    % full width
        before skip=0pt,     % remove space before box
        after skip=0pt,      % remove space after box
        left=5pt,           % reduce left margin
        right=4pt,          % reduce right margin
        title=Relationships Extraction Prompt
]     % allow page breaks

{\scriptsize  % or \small for slightly larger, \scriptsize for smaller
\begin{verbatim}
[INST] You are a software to extract relationships from sentences. 
Extract explicit and factual relationships between objects in the last sentence. 
Use the same formatting as below. No other text. 
One instance per subject, object, and predicate. Be exhaustive.

Sentence: 'A video of a person on the  side of a table holding food.'
subject: person, predicate: on the side of, object: table
subject: person, predicate: holding, object: food

Sentence: 'A kid touching the table  while sitting on a chair.'
subject: kid, predicate: touching, object: table 
subject: kid, predicate: sitting on, object: chair

Sentence: 'A man putting on shoes and clothes. 
Behind him two trees next to each other.'
subject: man, predicate: holding, object: shoe
subject: man, predicate: holding, object: clothes
subject: two trees, predicate: behind, object: him
subject: tree, predicate: next to, object: tree

Sentence: 'Woman sets table with plates, silverware, glasses, 
before placing oatmeal pot and juice pitcher in center. Calls family.'
subject: woman, predicate: set, object: table
subject: woman, predicate: set, object: plates
subject: woman, predicate: set, object: silverware
subject: woman, predicate: set, object: glasses
subject: woman, predicate: placing, object: oatmeal pot
subject: woman, predicate: placing, object: juice pitcher
subject: oatmeal pot, predicate: in center of, object: table
subject: juice pitcher, predicate: in center of, object: table
subject: woman, predicate: call, object: family

Sentence: 'Children playing on swings and slide. Couple sits on bench, 
holding hands.'
subject: children, predicate: playing on, object: swings
subject: couple, predicate: sit on, object: bench
subject: couple, predicate: holding, object: hands [/INST]
Sentence: {sentence}
\end{verbatim}
}
\end{tcolorbox}

\vspace{5mm}
\caption{Prompt for Extracting Relationships from Sentences}
\label{fig:relationship_extraction}
\end{figure*}

\end{document}